\documentclass[letterpaper, 10 pt, conference]{ieeeconf}
\IEEEoverridecommandlockouts
\usepackage{cite}
\usepackage{amsmath,amssymb,amsfonts}
\usepackage{algorithmic}
\usepackage{graphicx}
\usepackage{textcomp}
\usepackage{xcolor}

\usepackage{multicol}
\usepackage[bookmarks=true]{hyperref}
\usepackage{caption}
\usepackage{subcaption}
\usepackage{stfloats}
\usepackage{booktabs}
\usepackage{makecell}
\usepackage{multirow}
\usepackage{color}

\hypersetup{hidelinks}

\newcommand{\algo}{{\small \sc\textsf{TSHT}}}
\newcommand{\model}{{\small \sc\textsf{MS-Human-700}}}

\def\BibTeX{{\rm B\kern-.05em{\sc i\kern-.025em b}\kern-.08em
    T\kern-.1667em\lower.7ex\hbox{E}\kern-.125emX}}

\title{\LARGE \bf Self Model for Embodied Intelligence: \\ Modeling Full-Body Human Musculoskeletal System and Locomotion Control with Hierarchical Low-Dimensional Representation}

\author{Chenhui Zuo$^{*}$, Kaibo He$^{*}$, Jing Shao, Yanan Sui
\thanks{*These authors contributed equally to this work.}
\thanks{The authors are with the National Engineering Research Center of Neuromodulation, School of Aerospace Engineering, Tsinghua University, Beijing, China 100084. Correspondence to Yanan Sui. E-mail: \texttt{ysui@tsinghua.edu.cn}}
}

\begin{document}

\maketitle

\begin{abstract}

Modeling and control of the human musculoskeletal system is important for understanding human motor functions, developing embodied intelligence, and optimizing human-robot interaction systems. However, current human musculoskeletal models are restricted to a limited range of body parts and often with a reduced number of muscles. There is also a lack of algorithms capable of controlling over 600 muscles to generate reasonable human movements. To fill this gap, we build a musculoskeletal model (\model) with 90 body segments, 206 joints, and 700 muscle-tendon units, allowing simulation of full-body dynamics and interaction with various devices. We develop a new algorithm using low-dimensional representation and hierarchical deep reinforcement learning to achieve state-of-the-art full-body control. We validate the effectiveness of our model and algorithm in simulations with real human locomotion data. The musculoskeletal model, along with its control algorithm, will be made available to the research community to promote a deeper understanding of human motion control and better design of interactive robots.

Project page: \href{https://lnsgroup.cc/research/MS-Human-700}{https://lnsgroup.cc/research/MS-Human-700}

\end{abstract}

\section{Introduction}
\label{sec:intro}

The human musculoskeletal system is a complex dynamic system with high spatio-temporal degrees of freedom. While current technological advancements cannot effectively measure the dynamic characteristics of the human body as a whole, modeling and simulation are very important for the understanding of human motion control and human factors in human-robot interaction. Dynamic modeling can also facilitate the design of humanoid robots and provide a self-model of human for embodied intelligence.
A musculoskeletal model is built from anatomical and biomechanical data of the human body. By simulating human motion with a musculoskeletal model, it is possible to gain insight into the dynamics of the human body and the factors that influence motion, such as muscle activation patterns, joint stability, and limb coordination. 
We can use simulation data instead of hard-to-obtain real data to design and optimize robots, such as humanoids, exoskeletons, prosthetics, and other human-interactive devices.


Many of our daily movements require the use of the entire body. For instance, walking and running involve the coordinated movement of the lower limbs for propulsion, complemented by the swinging motion of the upper limbs for balance, while the trunk muscles contribute to torso stability. However, there is no musculoskeletal model that represents the entire human body. Constructing a full-body human musculoskeletal model enables comprehensive characterization of the action space and state space in human movement simulations. The dynamics of the model improves our understanding of human motor intelligence. Furthermore, current algorithms struggle to address the highly dimensional control problem associated with whole-body human musculoskeletal motion.

The modeling and control of the musculoskeletal model of full human body is profoundly difficult due to the following challenges. First, building a human musculoskeletal model requires detailed quantitative knowledge of anatomy and physiology. Second, the muscle control parameter space is complex, where over 600 skeletal muscles control hundreds of joints, leading to high dimensionality and redundancy of the system\cite{ting2012review, bernstein1966co}. Third, the dynamical property of neuro-muscle actuators is non-linear and inconstant\cite{zajac1989muscle, wolpert2000computational}. 

\begin{figure}[t]
  \centering
    \subcaptionbox{\label{fig:model_a}}
    {\includegraphics[width=0.3\linewidth]{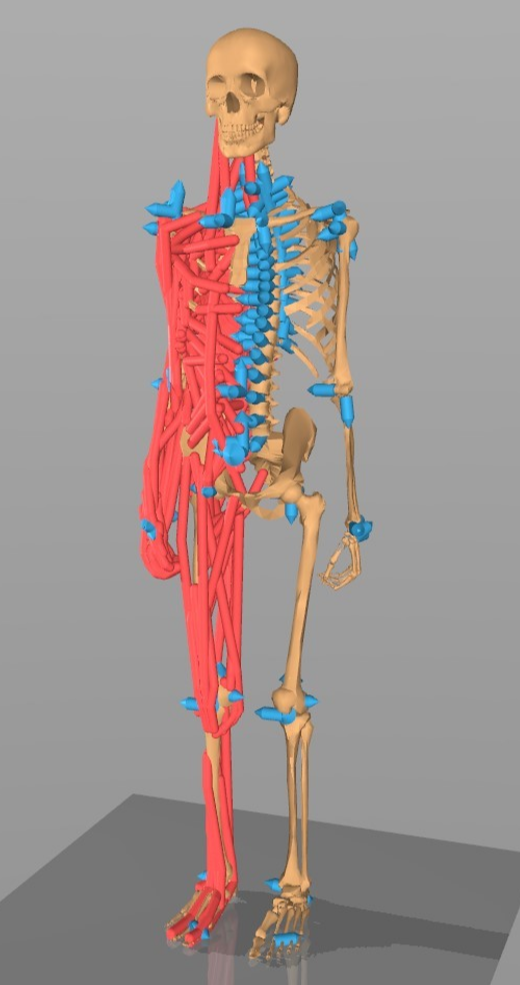}}
    \subcaptionbox{\label{fig:model_b}}
    {\includegraphics[width=0.3\linewidth]{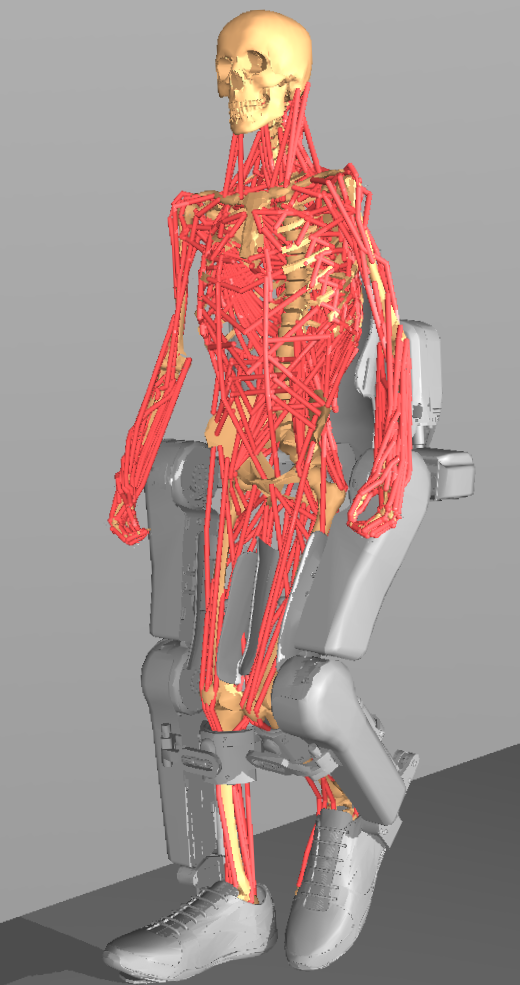}}
    \subcaptionbox{\label{fig:model_c}}
    {\includegraphics[width=0.3\linewidth]{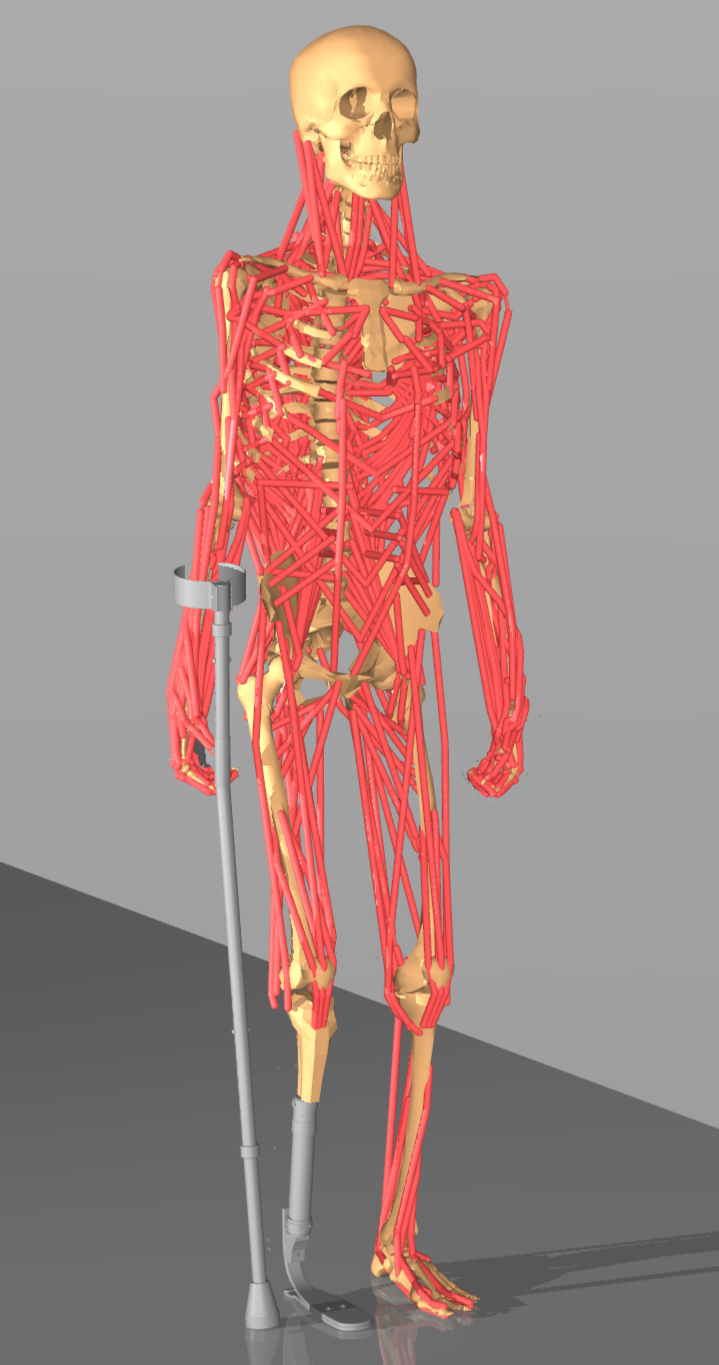}}
  \caption{(a) \model: full-body musculoskeletal model. Blue arrows represent joint axes, red lines represent muscle-tendon units (right half). (b) The model interacting with an exoskeleton in simulation. (c) The model with a prosthetic leg and a crutch.}
  \label{fig:fullbodyModels}
\end{figure}

Here, we develop \model, a full-body MusculoSkeletal Human model with 90 body segments, 206 joints, and 700 muscle-tendon units for neuromuscular-driven dynamic simulation, as shown in Fig.\ref{fig:fullbodyModels}. The model can be imported into open-source physics simulation engines like MuJoCo\cite{todorov2012mujoco} and OpenSim\cite{delp2007opensim}, establishing a framework that can be used to perform neuro-musculo-skeletal control simulation of human full-body movements.

We also develop a hierarchical deep reinforcement learning (RL) algorithm with a low-dimensional representation, Two-Stage Hierarchical Training (\algo), which is capable of controlling the model to generate biologically plausible movements. We make both the musculoskeletal model and our control algorithm available to the research community with the goal of fostering the development of more accurate models, the design of better assistive robots, and a better understanding of human neuromuscular motion control for embodied intelligence.


\textbf{Our Contributions.} 
We build a dynamic model of full-body human musculoskeletal system with physiologically feasible body segments, joints, and multi-muscle-tendon units.
We embed the whole-body model in simulation environments that support contact interaction simulations, providing a scalable, adaptable, and computationally efficient simulation framework for data-driven musculoskeletal system control.
We propose a deep reinforcement learning algorithm, two-stage hierarchical training, for the control of the high-dimensional and highly nonlinear full-body musculoskeletal model. Our control algorithm can generate desired motions in full-body musculoskeletal model.

\section{Related Work}

\begin{figure*}[t]
  \centering
  \includegraphics[width=0.9\linewidth]{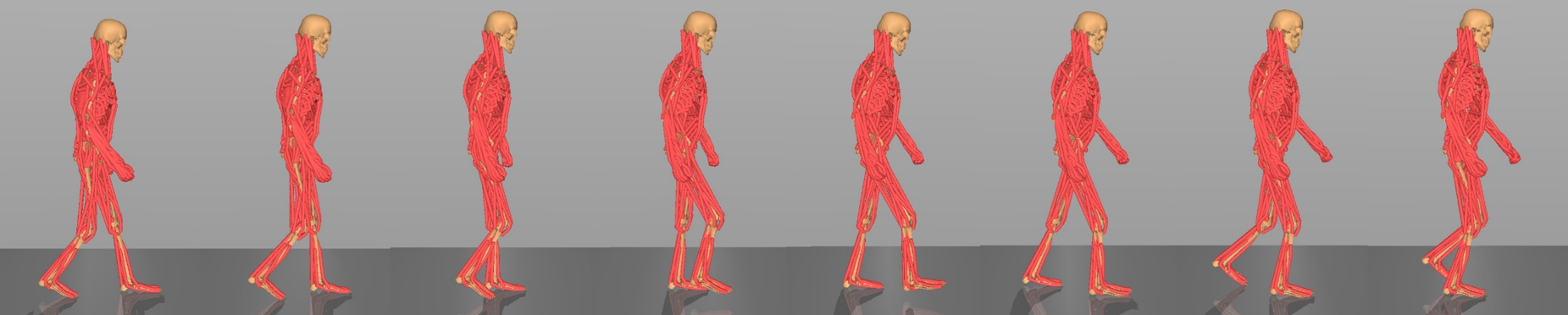}
  \caption{Learning to walk with \model~via Two-Stage Hierarchical Training (\algo).}
  \label{fig:gait}
\end{figure*}

\textbf{Human Musculoskeleton Modeling.} Many previous studies focused on the musculoskeletal models actuated by Hill-type muscles\cite{zajac1989muscle} for specific human body parts. An upper extremity model has been built and validated\cite{lee2009comprehensive, saul2015benchmarking, mcfarland2019spatial}. Lower extremity models have been continuously improved and used in the field\cite{delp1990interactive, arnold2010model, rajagopal2016full, sui2017quantifying,lai2017antagonist}. Some researchers built the model of the trunk\cite{bruno2015development} and extended it with simple muscles-actuated limbs\cite{schmid2020musculoskeletal}. Other researchers created models with muscles of the torso and limbs, but the models do not cover a large range of important skeletal muscles and joints of the torso\cite{lee2019scalable, chang2019full}. These models were mainly built on physics-based musculoskeletal simulation platforms. Representative research efforts include OpenSim \cite{delp2007opensim}, Anybody\cite{damsgaard2006analysis}, etc. However, running dynamic simulations on them is computationally expensive. Some recently developed physics engines\cite{todorov2012mujoco, coumans2016pybullet, lee2018dart, hwangbo2018per} are becoming popular for robotic simulation, and they are capable for efficient dynamic simulations that support contact interactions. MuJoCo's modeling approach is able to provide a simplified and fully supported interface for muscle modeling among them. Efforts have been made to convert the musculoskeletal models from OpenSim to MuJoCo, resulting in acceleration of simulation speed on the model by two orders of magnitude\cite{ikkala2020converting, wang2022myosim, MyoSuite2022}.

\textbf{Motion Control of the Musculoskeletal System.} Researchers have explored the use of biological actuators for controlling human musculoskeletal systems. Evolutionary algorithms were used to optimize control parameters for a 2D lower limb model comprising 16 muscle-tendon units\cite{wang2012optimizing}, and previous work \cite{10.1145/2661229.2661233} developed a more complex musculoskeletal model. However, these methods fell short when handling complex models. Recently, a hierarchical RL algorithm, coupled with imitation learning, was applied to a musculoskeletal model featuring 346 muscles\cite{lee2019scalable}. Xu et al.\cite{https://doi.org/10.48550/arxiv.2204.07137} introduced a high-performance differentiable simulator and an algorithm that effectively enhanced control over the human musculoskeletal model by leveraging gradient information. A review was conducted on the application of RL in neuromechanical simulations\cite{song2021deep}. Beyond the human musculoskeletal model, a 3D musculoskeletal model of an ostrich was constructed using the MuJoCo engine\cite{https://doi.org/10.48550/arxiv.2112.06061}. This model successfully performed various locomotion tasks with deep RL algorithm TD4. Schumacher et al.\cite{https://doi.org/10.48550/arxiv.2206.00484} proposed the DEP-RL algorithm to attain superior control over the aforementioned ostrich model. Furthermore, the concept of reward shaping to achieve natural walking in musculoskeletal models of human lower limbs was introduced\cite{weng_natural_2021}. Multi-task learning\cite{caggiano2023myodex} and synergistic action representation\cite{cheng2019motor, berg2023sar} have demonstrated their effectiveness in the motion control of specific musculoskeletal body parts.


In the open-source musculoskeletal modeling community, the current model with the largest number of muscles contains 552 individual muscle units\cite{schmid2020musculoskeletal}, which lacks the coverage of limb muscles to complete the simulation of full-body movements. Control algorithms that can drive the high-dimensional musculoskeletal models to produce realistic motions are still lacking.  
\section{Musculoskeletal System}

As demonstrated in Fig.\ref{fig:fullbodyModels}, our full-body musculoskeletal model includes 90 rigid body segments, 206 joints, and 700 muscle-tendon units. The model can simulate the joint movements of the torso and limbs driven by skeletal muscles. We use musculoskeletal models\cite{saul2015benchmarking, mcfarland2019spatial, rajagopal2016full, lai2017antagonist, bruno2015development} on OpenSim platform as the references to set anatomical parameters of the following models of body segments, joints and muscle-tendon units, following the pipeline of \cite{ikkala2020converting}. We also embed the \href{https://gitlab.com/project-march/march}{MARCH} exoskeleton model in simulation to demonstrate human-machine interactive walking. 

\subsection{Skeletal Modeling}

We choose the pelvis as the root of the kinematic tree. We set the position, mass and inertia properties of each body segment to enable dynamic simulation of the model. The shape of each body segment is defined with mesh files, reserving their convex hulls for collision calculation during simulation. The body segments are connected mainly by hinge joints. There are also slide joints to represent translational characteristic between segments. The axis and range of motion of each joint are defined in the model. The coupling relationship between joints is modeled by polynomial equations\cite{walker1988effects,rajagopal2016full}.

The dynamics of the human musculoskeletal system can be formulated with the Euler-Lagrangian equations as
\begin{equation}
    M(q)\Ddot{q} + c(q,\Dot{q}) = J_{m}^T f_m(act) + J_c^T f_c + \tau_{ext},
\end{equation}
where $q$ represents the generalized coordinates of joints, $M(q)$ is the mass distribution matrix, $c(q,\Dot{q})$ is Coriolis and gravitational forces applied to the generalized coordinates. $f_m(act)$ is the vector representing muscle forces generated by all muscle-tendon units, and is determined by muscle activations ($act$). $f_{c}$ is the constraint force and torque, and $\tau_{ext}$ is the external force and torque from the interaction with environments. $J_{m}$ and $J_{c}$ are Jacobian matrices that map force and torque to the space of generalized coordinates. 

\subsection{Muscle Modeling}

Our full-body musculoskeletal model has 700 muscle-tendon units. They exert tensile forces on bones to move joints.

\begin{figure}[ht]
  \centering
    \subcaptionbox{\label{fig:sim_a}}
    {\includegraphics[width=0.4\linewidth]{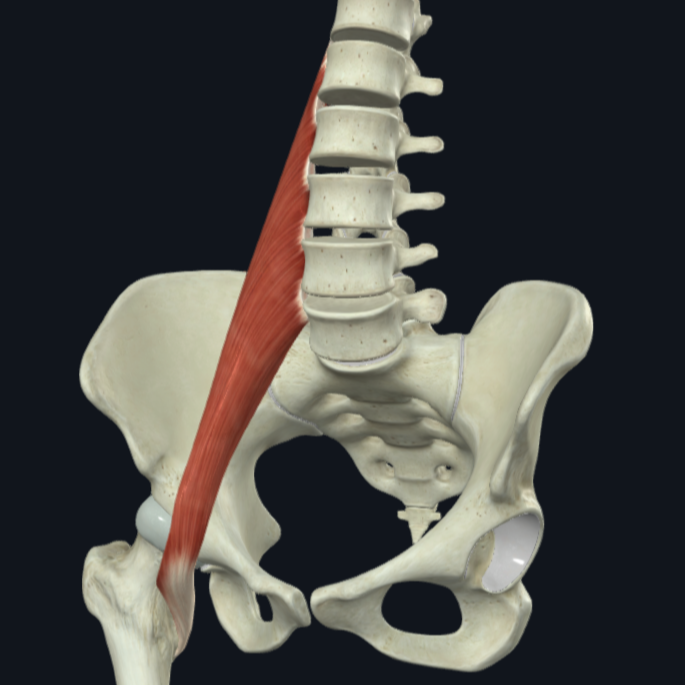}}
    \subcaptionbox{\label{fig:sim_b}}
    {\includegraphics[width=0.4\linewidth]{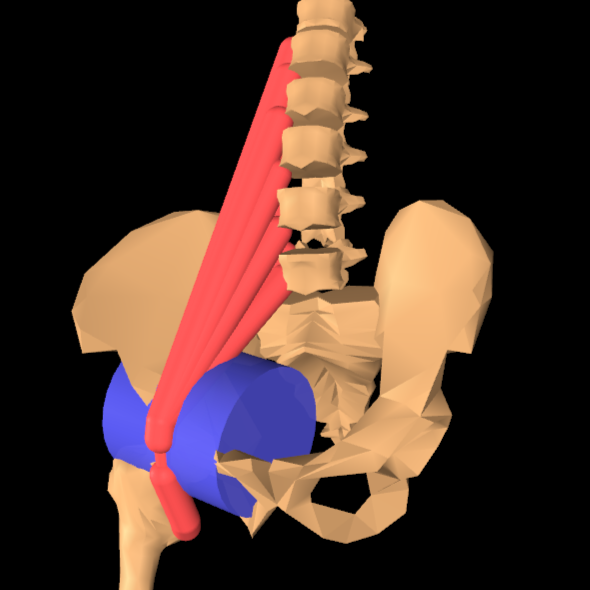}}
  \caption{(a) The anatomy of Psoas Major muscle from the medical atlas, image courtesy of \href{https://3d4medical.com/}{Complete Anatomy}. (b) Its multi-muscle-tendon units and the wrapping geometry constructed in our model. }
  \label{fig:simulation}
\end{figure}

\textbf{Muscle-Tendon Pathway.} These muscle-tendon units typically originate and are inserted at pathway points attached to different body segments. For wide muscle attachments, it will be modeled as multiple independent muscle-tendon units. Thus, the total number of units, 700, in our model is larger than the common biomedical estimations for the number of skeletal muscles, which is around 640 \cite{gray2013gray}. Wrapping geometry surfaces are added to the pathways to model physical constraints imposed by bones and soft tissue when needed. Our multi-muscle-tendon and wrapping geometry models are demonstrated in Fig. \ref{fig:sim_a} and Fig. \ref{fig:sim_b}.


\textbf{Neuro-Muscle Dynamics.} The activation-contraction dynamics of muscles exhibit non-linearity and temporal delay, thereby posing challenges to neuromuscular control. With the employment of the Hill-type muscle model\cite{zajac1989muscle}, the maximum isometric force, the tendon length, and the optimal fiber length of each unit is defined in the model. The force produced by a single muscle-tendon unit can be formulated as 
\begin{equation}
f_m(act)=f_{max}\cdot [F_{l}(l_m)\cdot F_v(v_m)\cdot act + F_p(l_m)],
\end{equation}
where $f_{max}$ stands for the maximum isometric muscle force and $act, l_m, v_m$ respectively stand for the activation, normalized length and normalized velocity of the muscle. $F_l$ and $F_v$ represent force-length and force-velocity functions fitted using data from biomechanical experiments\cite{millard2013flexing}. The muscle activation $act$ is calculated by a first-order nonlinear filter in simulation as Eq. (\ref{equ:muscle_dynamic}) where $u$ is the input control signal of the musculoskeletal model. $(\tau_{\text{act}}, \tau_{\text{deact}})$ is a time constant for activate and deactivate latency with defaults $(10ms, 40ms)$. 
\begin{equation}
    \frac{\partial act}{\partial t} = \frac{u-act}{\tau(u,act)},
    \tau(u,act) = \begin{cases}\tau_{\text{act}}(0.5+1.5act) & u>act\\

    \frac{\tau_{\text{deact}}}{0.5+1.5act} & u\leq act\end{cases}
    \label{equ:muscle_dynamic}
\end{equation}


\section{Control}

To tackle this high-dimensional musculoskeletal model control problem, we propose a training algorithm based on hierarchical action space representation, which is capable of generating natural motions through two stages of training.

\begin{figure*}[ht]
  \centering
  \includegraphics[width=0.85\linewidth]{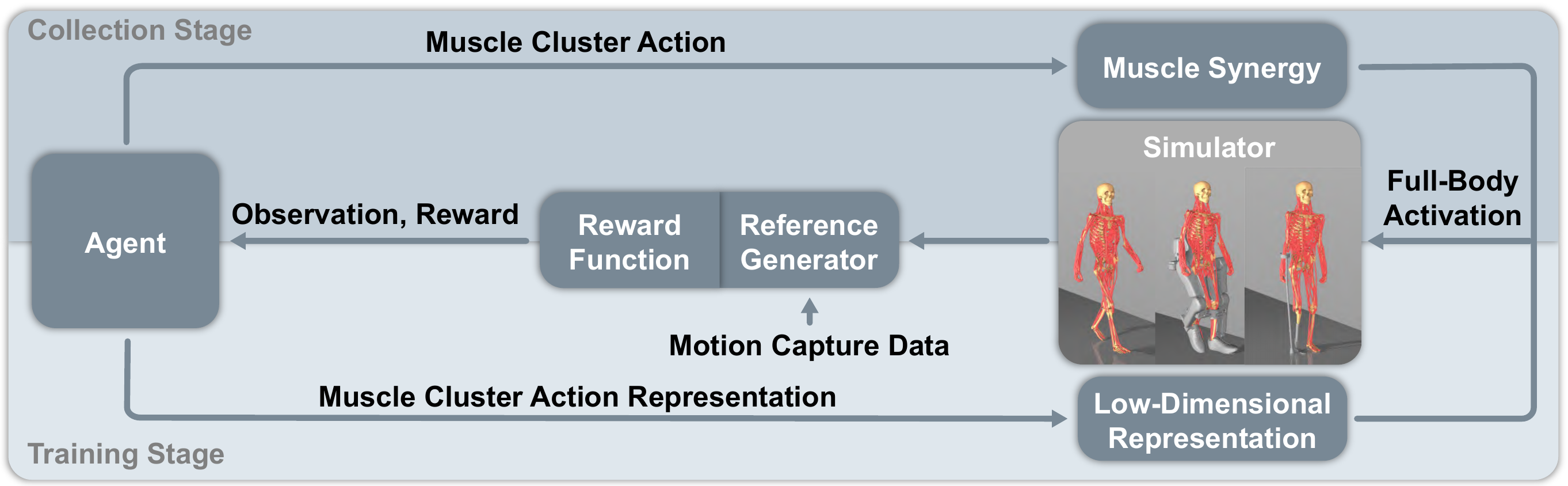}
  \caption{Two-Stage Hierarchical Training (\algo). In the collection stage, the agent determines muscle cluster actions. Muscle Synergy function then maps these muscle cluster actions into full-body muscle activation, leveraging prior physiological knowledge. Meanwhile, the reference generator generates the desired joint states based on motion capture data and computes corresponding rewards. Once a sufficient number of sub-optimal trajectories are collected, low-dimensional representations are extracted by encoder-decoder architecture and utilized for the training stage.}
  \label{fig:algo}
\end{figure*}

\subsection{Problem formulation for Musculoskeletal Model Control}

We formulate the control problem of the musculoskeletal model as a Markov Decision Processes, denoted by $ \langle \mathcal{S}, \mathcal{A}, r, p, \rho_0 , \gamma \rangle $, where $\mathcal{S} \subseteq \mathbb{R}^n$ represents the continuous space of all valid states, $\mathcal{A} \subseteq \mathbb{R}^m$ represents the continuous space of all valid actions. $r: \mathcal{S} \times \mathcal{A} \to \mathbb{R}$ is the reward funtion. $p$ stands for the state transfer probability function, where $ p(s'|s,a)$ denotes the probability of an agent taking an action $a$ to transfer from the current state $s$ to the state $s'$, $\rho_0$ is the probability distribution of the initial state with $\sum_{s\in \mathcal{S}} \rho_0(s)=1$ and $\gamma \in [0, 1)$ is the discount factor. We adopt an off-policy RL algorithm, Soft Actor-Critic (SAC)\cite{haarnoja2018soft}, to optimize our policy. This algorithm aims to obtain the optimal policy by maximizing the accumulative discounted return, $\sum_t \gamma^t R_t$ and the policy entropy, $\mathcal{H}(\pi)$, where
\begin{equation}
    \pi^*=\arg \max _\pi \mathbb{E}_{(o_t, a_t)} \left[\sum_{t=0}^{T} \gamma^t\left(r_t+\alpha \mathcal{H}\left(\pi\left(\cdot \mid s_t\right)\right)\right)\right].
\end{equation}

State space $\mathcal{S}$ consists of $\{t, q, \dot q, f_m, l_m, \dot l_m , act, \mathcal{E} \}$, represents simulation time, joint position, joint velocity, muscle force, muscle length, muscle velocity, muscle activation, and other task-specific observations respectively. Task-specific observations are introduced in section \ref{Exp_Setup}. Action space $\mathcal{A}$ consists of muscle excitation $u$. We can also call muscle excitation as motor neuron signals.

\subsection{Motion Imitation}
\label{sec:motionimitaion}


When the reward function is defined in a straightforward manner using forward velocity, the agent is still able to learn a policy to get a high reward, but the gait turns out to be very unnatural. Given the complex nature of elaborate human musculoskeletal models, researchers commonly regulate the search space by reward shaping. Using tracking motion capture clips is one of the very effective reward setting schemes. A recent work\cite{https://doi.org/10.48550/arxiv.2112.06061} integrated motion capture data into the reward function to generate natural movements.

The reference generator shown in Fig. \ref{fig:algo} can generate reference joint position $q^r$ and velocity $\dot{q}^r$ taking motion capture data from different tasks. The optical motion capture system calculated the position of markers, based on which we perform an inverse kinematic to reconstruct motions. We then obtain a time series of joint angles. Taking the distance between the actual joint state $\left \{q, \dot{q} \right \}$ and the desired joint state $\left \{q^r, \dot{q}^r\right \}$ as input, the policy is able to accomplish all considered tasks efficiently.

\subsection{Hierarchical Action Space Representation}

Due to the high-dimensional nature of the action space, previous algorithms\cite{https://doi.org/10.48550/arxiv.2112.06061, https://doi.org/10.48550/arxiv.2206.00484} struggled to maintain control over our full-body musculoskeletal model during locomotion generation. Consequently, the inadequately controlled models frequently collapsed during simulation. Given the vastness of the action space, resolving this issue hinges on the representation of the action space. We employ physiological muscle synergy and a low-dimensional representation of muscle activation to expedite the policy learning process.

\textbf{Physiological Muscle Synergy.} We optimize our policy according to the muscle synergy effect \cite{chvatal2013common,ting2007neuromechanics,d2003combinations}. Components in human musculoskeletal system are highly redundant. However, each of the muscles is not activated individually. Muscles work synergistically or antagonistically during human movement, where many of them often form clusters that are being activated or deactivated simultaneously for motor control. Muscle synergy could contribute to dimension reduction. In our experiments, we divide the muscles into anatomical clusters. Muscle Synergy (MS) function assign the excitation to each muscle-tendon unit in the muscle clusters. The control signal, muscle excitation $u$, is determined by the muscle cluster action $u_c$, where $u = f_c(u_c)$. $f_c$ is the function mapping the cluster action space into the muscle excitation space. Due to the high degree of freedom of muscles in the torso, we introduced biological muscle synergy. For limb muscles, the control of each muscle-tendon unit is independent.

\textbf{Low-Dimensional Representation of Muscle Activation.} From a physiological point of view, the majority of high-dimensional human motion tasks are controlled by low-dimensional signals from the brain. Control by relying solely on physiological muscle synergy often fails to perform very well and accurately. Therefore, how to find potential representations in the temporal signals of a large number of muscle activations is the key to solving the problem. Manifold learning\cite{izenman2012introduction} aims to find a low-dimensional representation in high-dimensional data as the essential features. AutoEncoder is a powerful neural network architecture that efficiently learns low-dimensional representations from the data. Works such as VAE\cite{kingma2013auto} utilize a large amount of data to find the distribution that fulfills the image generation task. In our work, we use a encoder $E_\phi: \mathcal{X} \in \mathbb{R}^n \to \mathcal{Z} \in \mathbb{R}^d$ that embed the high-dimensional data into a latent space, where the decoder $D_\phi: \mathcal{Z} \in \mathbb{R}^d \to \mathcal{X} \in \mathbb{R}^n$ can be used to recover the original data. We use certain sub-optimal muscle-activating signals to formalize the encoder and the decoder. The agent is then trained in the latent space.

\subsection{Two Stages Hierarchical Policy Training}

Little effective information can be obtained by adopting End-to-End (E2E) strategies when controlling the motion of musculoskeletal models with imitation learning. Therefore, we divide the entire training process into two stages, namely the collection stage and the training stage. The entire training algorithm is shown in Fig. \ref{fig:algo}.

\textbf{Collection Stage.} Consider muscle synergy to improve the E2E methods to a certain extent, and then extract low-dimensional representations of the trajectories generated by this policy. At this stage, the policy controls the action of each muscle cluster, and the muscle synergy maps it back to the high-dimensional space of all muscle activations.

\textbf{Training Stage.}  Before attaining a detailed understanding of human motor control, people remain restricted by the inherent limitations of physiological muscle synergy. We therefore employ an encoder-decoder architecture to extract more precise low-dimensional representations from the generated trajectories. During the collection stage, we optimize the encoder-decoder architecture to minimize reconstruction errors. The encoder-decoder architecture encompasses various dimensionality reduction and expansion methods. In the training stage, we use the extracted low-dimensional representation for policy learning. The agent outputs the actions to be implemented in the low-dimensional space, and the control of the full-body model is completed through recovery realized by the decoder. By further fine-tuning the rewards, the trained agent can produce more natural actions.

\section{Experiments}

We compare the \algo~algorithm on \model~model with representative baselines, including SAR\cite{berg2023sar} and E2E RL. We design a series of full-body motion tasks and interaction scenarios to demonstrate the superiority of our model and algorithm.

\subsection{Experimental Setup} \label{Exp_Setup}

We designed three tasks: Walking forward, Human-Exoskeleton (Exo) interaction, and Prosthetic walking, to demonstrate full-body motion control, human-machine interaction, and further personalization of pathological conditions. In all environments, the joints on the torso are constrained to achieve better control, with the exception of certain spinal joints. We use optical motion capture to collect motion data for the motion imitation, as shown in Fig. \ref{fig:mocap_a}.

\textbf{Walking forward.} We expect the musculoskeletal model to mimic a realistic walking trajectory going forward. The trajectory is generated through the Reference Generator mentioned in section \ref{sec:motionimitaion}.

\begin{equation*}
    r =  10e^{-w_q(q-q^r)^2 - w_{\dot q}(\dot q - \dot q^r) - w_c(q_c - q_c^r)}
\end{equation*}

The reward consists of three parts, where $w_q$, $w_{\dot q}$ and $w_c$ are the weights of the reward for position tracking, velocity tracking and concerned joint position tracking respectively. $w_q = 1, w_{\dot q} = 5e^{-3}, w_c=1$. The musculoskeletal model may be initialized at any point throughout a trajectory cycle. The task does not have task-specific observation.

\textbf{Human-Exo interaction.} We also design a task on human interaction with the \href{https://gitlab.com/project-march/march}{MARCH} exoskeleton. In this environment, we expected the model to walk in the exoskeleton with minimal contact force in legs. Similarly, we use trajectory tracking for its upper limbs. The joint angles of the exoskeleton are tracking the reference trajectory by a PID controller. We use spring elements with high stiffness to model the bandage interaction between legs and exoskeleton.

\begin{equation*}
    r = w_m e^{-(q_u-q_u^r)^2} - w_c \left \| f_{\text{contact}  }\right \|_2
\end{equation*}

The reward comprises a penalty for contact force and a reward for tracking the upper limb trajectory, where $w_m$ and $w_c$ are the weights of the corresponding rewards. $w_m=10, w_c=5e^{-2}$. The task-specific observations for this task include the distance between the ends of the elastic elements on the leg and the exoskeleton.

\textbf{Prosthetic walking.} We replace the right calf and foot of the full-body model with a prosthetic and add a crutch to its right hand. This is done to demonstrate its resistance to external disturbances and imitate gait under pathological conditions. The reward and observations are identical to those used in the walking forward task. 

\subsection{Baselines}

We compared our algorithm with current state-of-the-art methods: E2E, SAR, and SAR-ours. Our algorithm begins with the collection stage over a time step of length $M$ with muscle synergy, followed by optimization over a training stage of length $N$. When selecting of encoder-decoder architecture, inspired by SAR, we employed PCA\cite{ringner2008principal} and ICA\cite{stone2004independent}, where $u_c=\text{ICA}(\text{PCA}(u))$. In contrast, E2E is directly trained over a time step of $M+N$. SAR collects low-dimensional representations during a time step of $M$ and is trained over a time step of $N$. SAR-ours, on the other hand, extracts low-dimensional representations using muscle synergy during the time step of $M$ and then trains over a time step of length $N$.

\subsection{Experimental Results}

\begin{figure}[htbp]
  \centering
  \subcaptionbox{\label{fig:performance_gait}}{\includegraphics[width=0.49\linewidth]{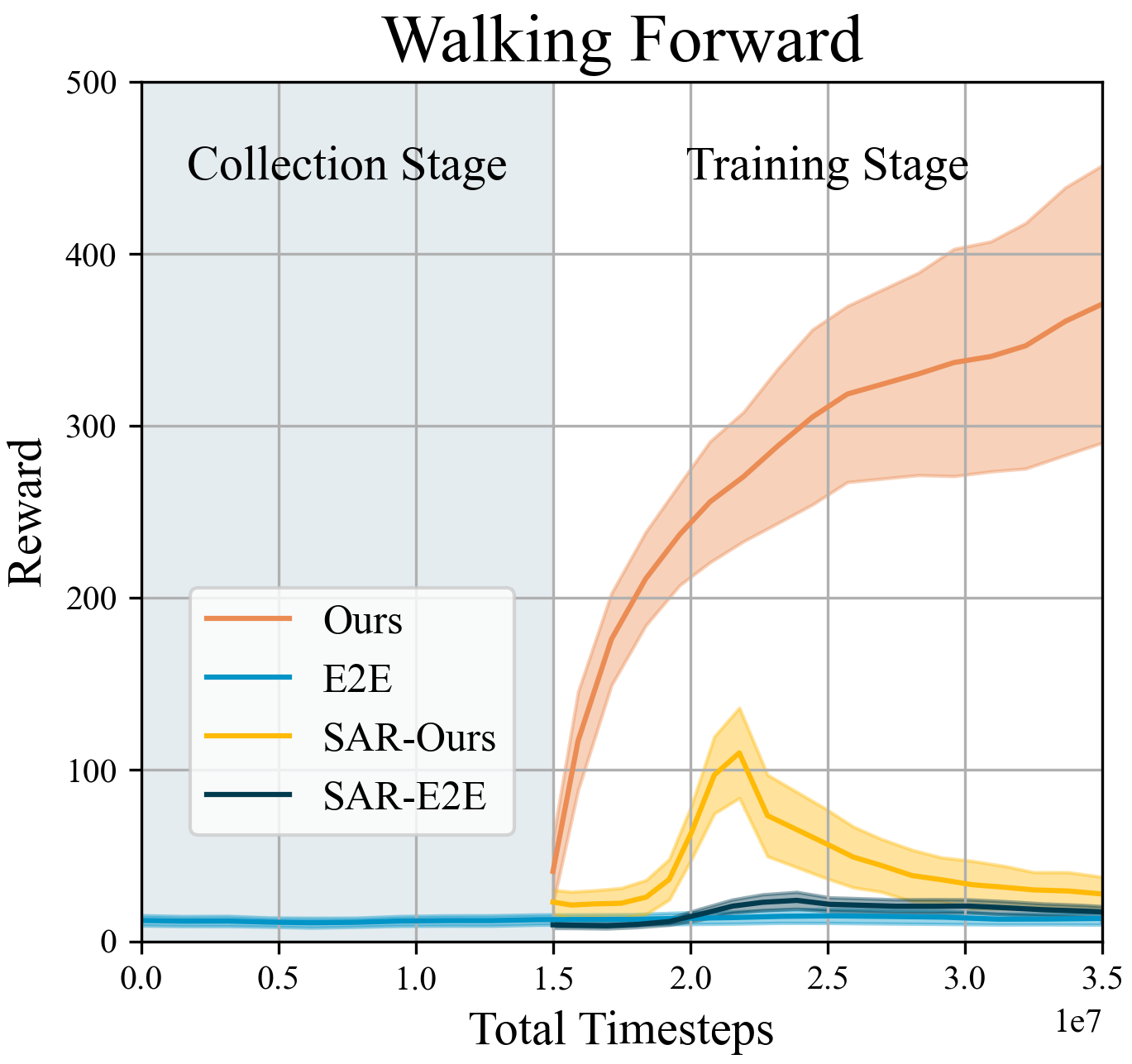}}
  \subcaptionbox{\label{fig:performance_exo}}{\includegraphics[width=0.49\linewidth]{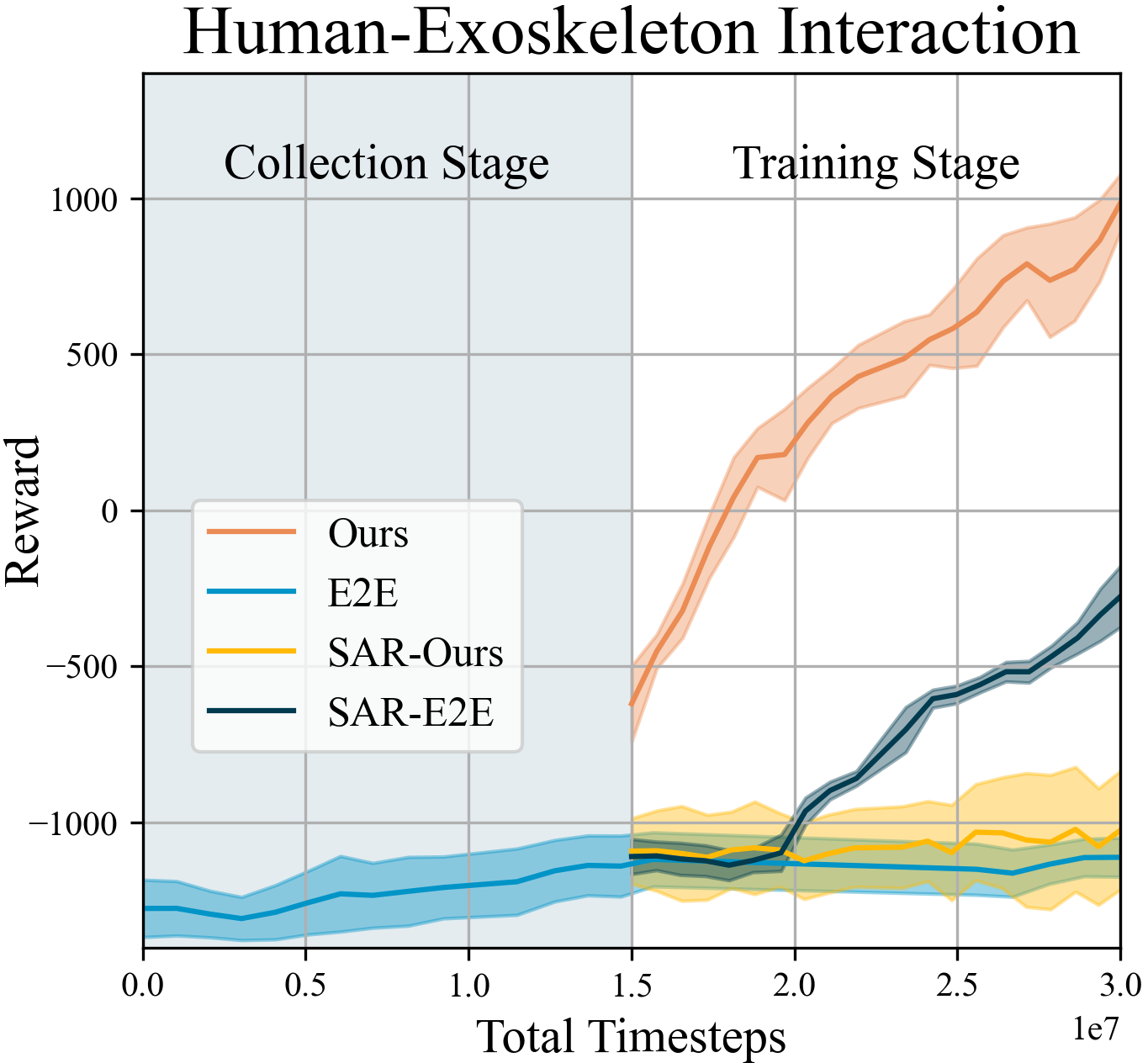}}
    \subcaptionbox{\label{fig:performance_pro}}{\includegraphics[width=0.49\linewidth]{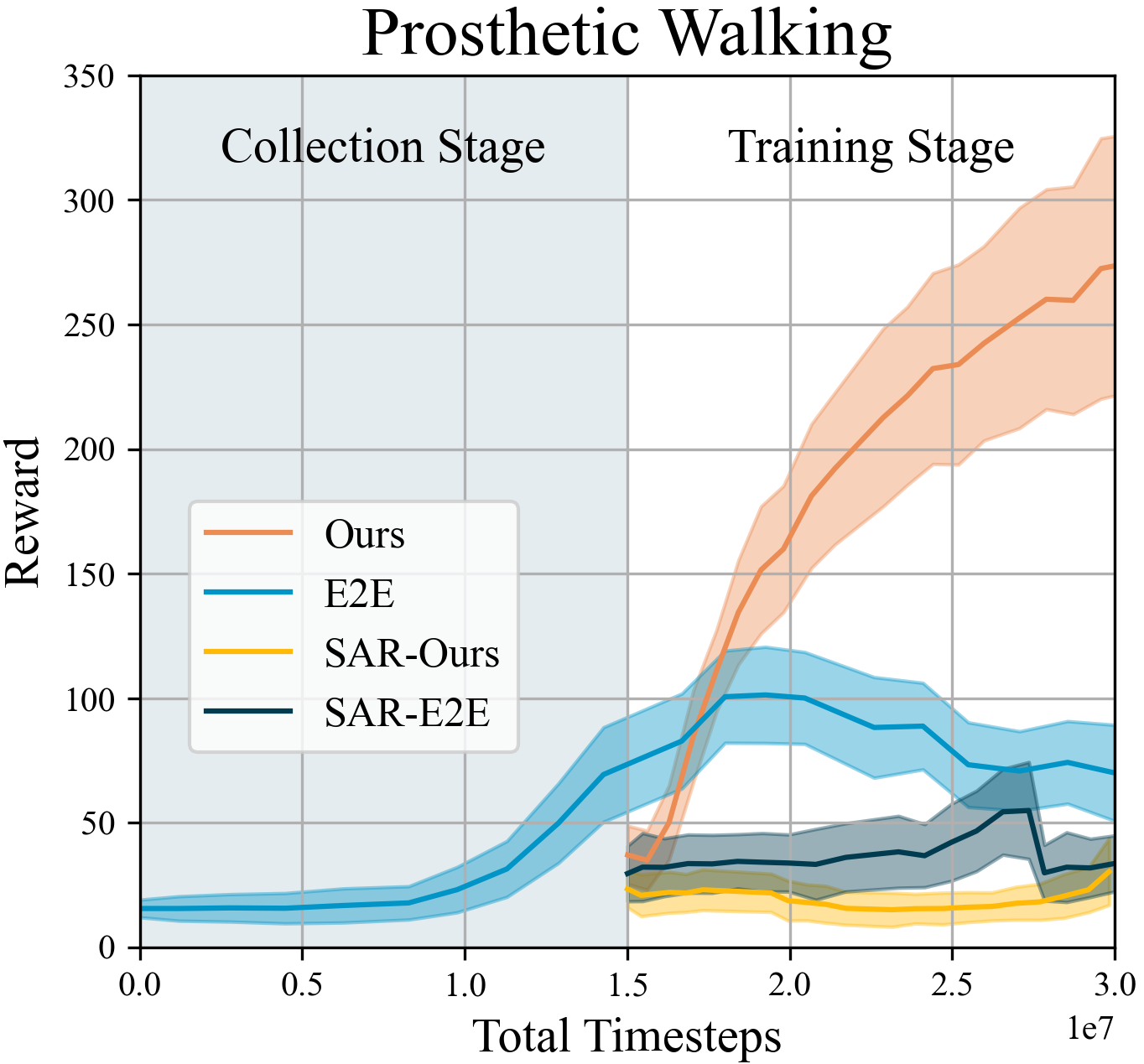}}
  \hspace{5mm}
  \subcaptionbox{\label{fig:result_pros}}{\includegraphics[width=0.27\linewidth]{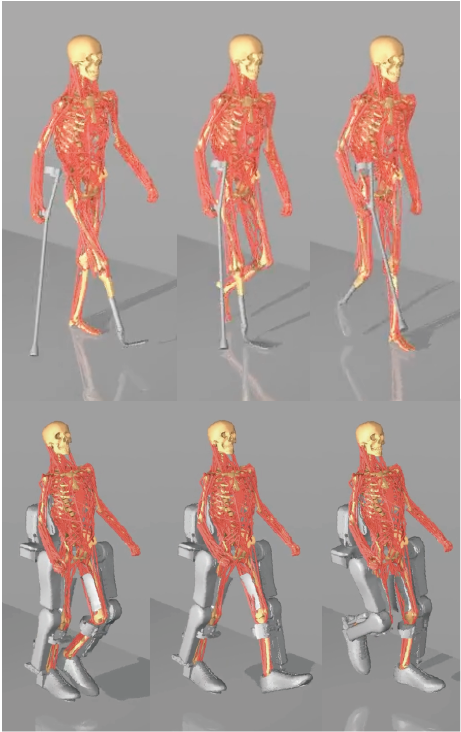}}
  \caption{(a)-(c) Our algorithm outperforms baseline algorithms on different tasks. (d) Learnt locomotion with a prosthetic leg and a crutch (top) and an exoskeleton (bottom).}
  \label{fig:result_curve}
\end{figure}

In our experiments, we used the SAC implementation from the open-source RL library stable-baselines3\cite{raffin2019stable}. For the Walking forward and Human-Exo interaction tasks, both the policy network and the value function network took dimensions of (512, 300). For Prosthetic walking, the dimensions were (400, 300), while all other parameters remained at their default settings. All observation values were normalized to a maximum value of 10. For the Walking forward task, the collection stage has a length ($M$) of 15M, while the training stage ($N$) is 20M. In the Human-Exo interaction and Prosthetic walking tasks, both $M$ and $N$ are set to 15M. The latent dimension $d$ is 50.


\begin{table}[htbp]
\caption{Comparing the maximum returns of algorithms on the three tasks.}
\centering
\begin{tabular}{@{}lcccc@{}}
\toprule
\multicolumn{1}{c}{}  & \textbf{Ours} & E2E & SAR & SAR-Ours \\ \midrule
Walking forward       & \textbf{371.07}&14.72&109.58& 23.78    \\
Human-Exo interaction &  \textbf{986.91}   &  -1111.48   &  -275.09   & -1022.01         \\
Prosthetic walking    &  \textbf{274.67}  & 101.24     &  54.83   &  30.28   \\ \bottomrule
\end{tabular}

\label{table:result}
\end{table}

By comparing our algorithm with baseline methods across all tasks, we achieved SOTA performance as shown in Table.\ref{table:result} and Fig. \ref{fig:result_curve}. Neither E2E nor SAR performs well in such high-dimensional tasks.

\begin{figure}[ht]
  \centering
    \subcaptionbox{\label{fig:mocap_a}}
    {\includegraphics[width=0.28\linewidth]{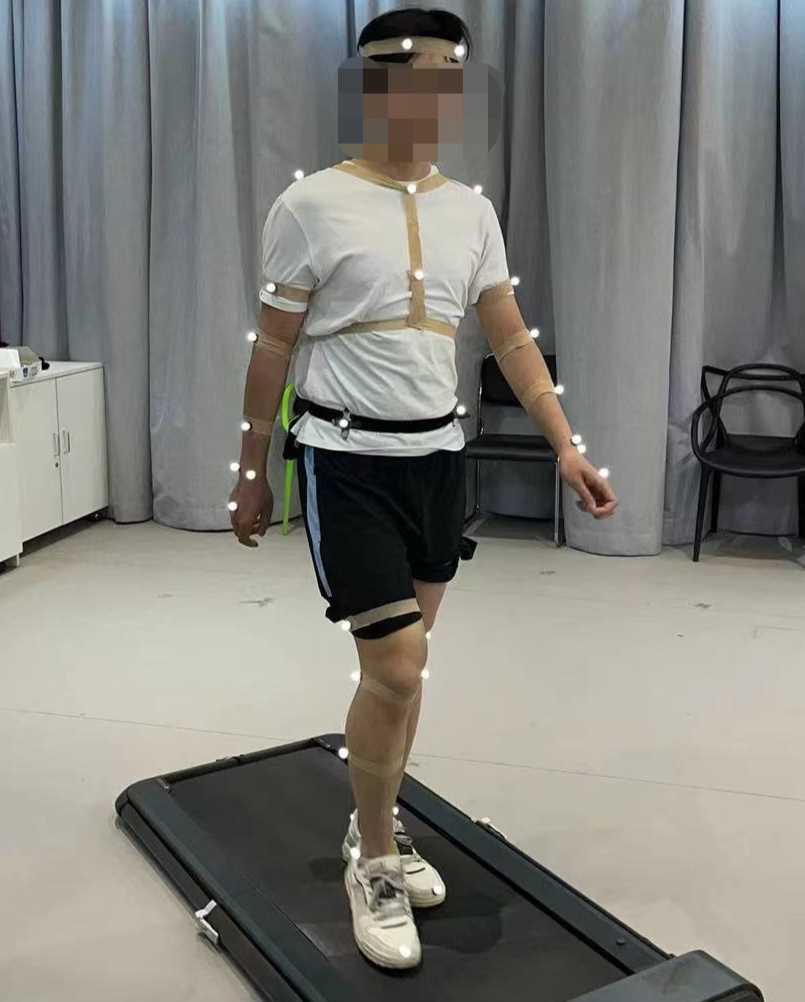}}
    \subcaptionbox{\label{fig:mocap_b}}
    {\includegraphics[width=0.27\linewidth]{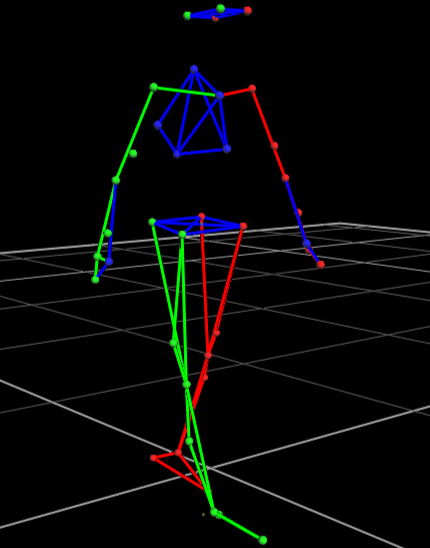}}
    
    \subcaptionbox{\label{fig:qpos}}
    {\includegraphics[width=0.35\linewidth]{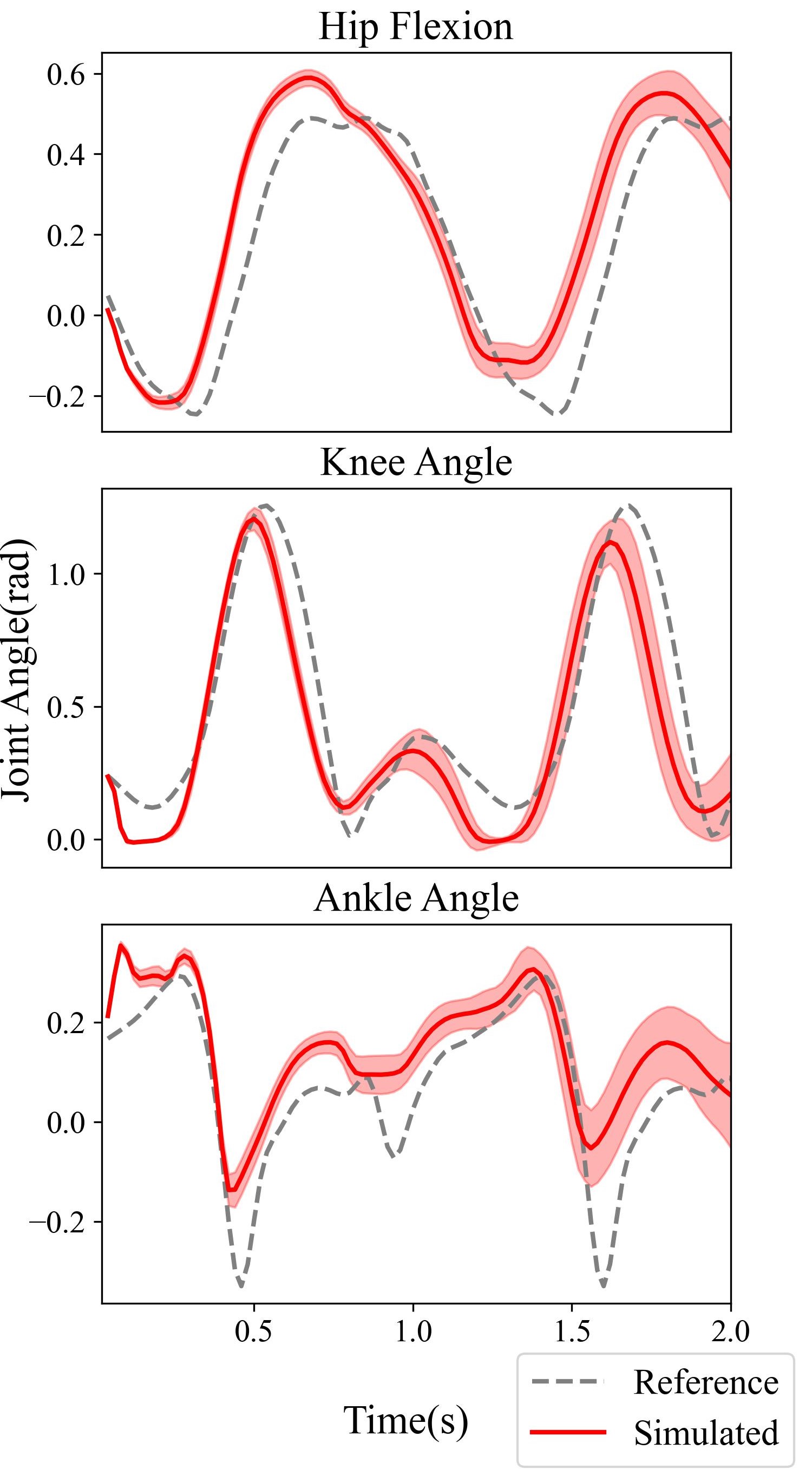}}
    \subcaptionbox{\label{fig:dtw_emg}}
    {\includegraphics[width=0.56\linewidth]{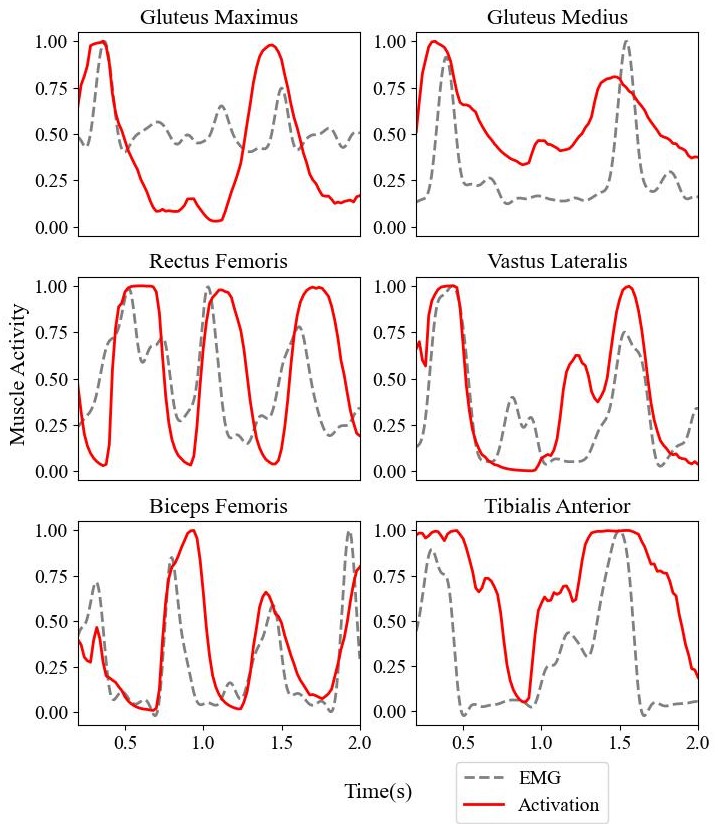}}
  \caption{(a) Our motion capture experiment of human walking. (b) The reconstruction of a full-body walking trajectory in \href{https://www.vicon.com/}{Vicon Nexus}. (c) The joint angles in the sagittal(X-Z) plane of the lower limbs, the positive joint directions indicates hip flexion, knee flexion and ankle dorsiflexion respectively. (d) The comparison of simulated muscle activations and the experimental EMG data in 6 representative muscle groups.}
  \label{fig:mocap_and_result}
\end{figure}

\subsection{Analysis and Discussion}

We demonstrate the visualization of our model's motion performance in Fig. \ref{fig:gait} and Fig. \ref{fig:result_curve}. We also provide a supplementary video. As shown in Fig. \ref{fig:result_curve}, in the human-exoskeleton interaction task, the algorithm is able to learn a policy that not only maintains upper body balance but also minimizes interaction forces between human and the exoskeleton.

Fig. \ref{fig:qpos} and Fig. \ref{fig:dtw_emg} shows the representative joints' angular position and muscle activity of the lower limbs in simulation. A walking simulation was presented along with the motion capture data and electromyography (EMG) recordings\cite{rajagopal2016full}, which we consider as our reference. The periodicity and stability of the curves substantiate the robustness of our algorithm's control efficacy. The discrepancy between the simulated and reference angular positions of the joints can be attributed to suboptimal reference trajectories that fail to ensure balance in the dynamics simulation of the full-body model. Consequently, an agent may opt for a more stable policy, such as increasing hip flexion and ankle dorsiflexion to elevate toe height and prevent tripping or falling, which is indicated in Fig. \ref{fig:qpos}. The relaxation training phase after imitation\cite{jin_high-speed_2022} may be used to improve the dynamic motion performance. The temporal pattern of simulated muscle activations corresponds with key features of the EMG signals recorded during walking, despite exhibiting discrepancies in timing and magnitude for certain muscle activations when compared to EMG data. Achieving simulated activations that accurately replicate the measured EMG activity across all muscles remains a crucial challenge in studies utilizing muscle-driven simulations\cite{rajagopal2016full, lee2019scalable}.

\section{Conclusion}



This work presents a musculoskeletal model (\model) with 90 rigid body segments, 206 joints, and 700 muscle-tendon units that encompasses the torso and limbs of human body. The model is adaptable to efficient open-source simulation engines, offering a simulation framework with a versatile interface for neuro-musculo-skeletal control simulation of human full-body movements. We introduce a novel control algorithm designed for high-dimensional and nonlinear musculoskeletal models to generate biologically plausible human motion. Our algorithm employs a hierarchical low-dimensional representation approach, integrating deep reinforcement learning and imitation techniques. Our proposed method exhibits superior performance in full-body musculoskeletal model control tasks, surpassing baseline algorithms significantly. Furthermore, the derived control policy of the agents provides valuable insights into the quantitative analysis of human motion control. Modeling and controlling musculoskeletal systems hold the potential to deepen our understanding of human motor intelligence and human factors in human-machine interactions. As a self-model of human for embodied intelligence, it could serve as a testing ground for the design of interactive robots and offer insights into humanoid behavior.

\vspace{12pt}


\bibliographystyle{IEEEtran}
\bibliography{IEEEabrv,References}

\end{document}